\documentclass[10pt,twocolumn,letterpaper]{article}

\usepackage{cvpr}
\usepackage{times}
\usepackage{epsfig}
\usepackage{graphicx}
\usepackage{amsmath}
\usepackage{amssymb}

\usepackage{multirow}
\usepackage{sidecap}
\usepackage{subfigure}
\usepackage{float}
\usepackage{amsmath}
\usepackage{textcomp}
\usepackage{booktabs}

\usepackage{tabularx}

\usepackage{graphicx}

\usepackage{float}
\usepackage{multirow}
\usepackage{diagbox}

\usepackage{cuted}
\usepackage{capt-of}
\usepackage{pifont}
\usepackage{color}

\usepackage[pagebackref=true,breaklinks=true,colorlinks,bookmarks=false]{hyperref}

\cvprfinalcopy 

\def\cvprPaperID{3977} 

\begin{document}

\title{Towards Fast and Accurate Real-World Depth Super-Resolution: Benchmark Dataset and Baseline}

\author{Lingzhi He, Hongguang Zhu, Feng Li, Huihui Bai, Runmin Cong,\\
	 Chunjie Zhang,	Chunyu Lin, Meiqin Liu, Yao Zhao\thanks{Corresponding author: yzhao@bjtu.edu.cn}\\
	Institute of Information Science, Beijing Jiaotong University\\
	Beijing Key Laboratory of Advanced Information Science and Network, Beijing, 100044, China\\
	{\tt\small \{lingzhihe, hongguang, l1feng, hhbai, rmcong, cjzhang, cylin, mqliu, yzhao\}@bjtu.edu.cn}

}

\maketitle
\pagestyle{empty}
\thispagestyle{empty}

\begin{abstract}
	Depth maps obtained by commercial depth sensors are always in low-resolution, making it difficult to be used in various computer vision tasks. Thus, depth map super-resolution (SR) is a practical and valuable task, which upscales the depth map into high-resolution (HR) space. However, limited by the lack of real-world paired low-resolution (LR) and HR depth maps, most existing methods use downsampling to obtain paired training samples. To this end, we first construct a large-scale dataset named ``RGB-D-D'', which can greatly promote the study of depth map SR and even more depth-related real-world tasks. The ``D-D'' in our dataset represents the paired LR and HR depth maps captured from mobile phone and Lucid Helios respectively ranging from indoor scenes to challenging outdoor scenes. Besides, we provide a fast depth map super-resolution (FDSR) baseline, in which the high-frequency component adaptively decomposed from RGB image to guide the depth map SR. Extensive experiments on existing public datasets demonstrate the effectiveness and efficiency of our network compared with the state-of-the-art methods. Moreover, for the real-world LR depth maps, our algorithm can produce more accurate HR depth maps with clearer boundaries and to some extent correct the depth value errors.
	
\end{abstract}

\section{Introduction}
As a supplement of the RGB modality, the depth map can provide useful depth information, which has been applied in bokeh rendering~\cite{Luo_2020_CVPR}, AR modeling~\cite{marchand2015pose}, face recognition~\cite{abate20072d}, gesture recognition~\cite{molchanov2015hand}, ~\etc. Meanwhile, the low-power depth sensors equipped on mobile consumer electronics, such as Huawei and Samsung, have been popular in our daily life. However, the resolution of depth maps cannot match the resolution of RGB images, limiting practical applications to some extent. Therefore, investigating the depth map SR is an effective solution for this issue. Furthermore, downsampling as a straightforward strategy has been widely used in the existing depth map SR algorithms~\cite{kim2020deformable,li2019joint} to construct paired training samples. But the downsampling manner fails to comprehensively simulate the real-world complex correspondences between the LR and HR depth maps. To bridge this gap, we construct the first benchmark dataset towards the real-world depth map SR. Furthermore, to meet the actual application requirements, we provide a fast and accurate depth map SR baseline model.
\begin{figure}[tbp]
	\begin{center}
		\includegraphics[width=1\linewidth]{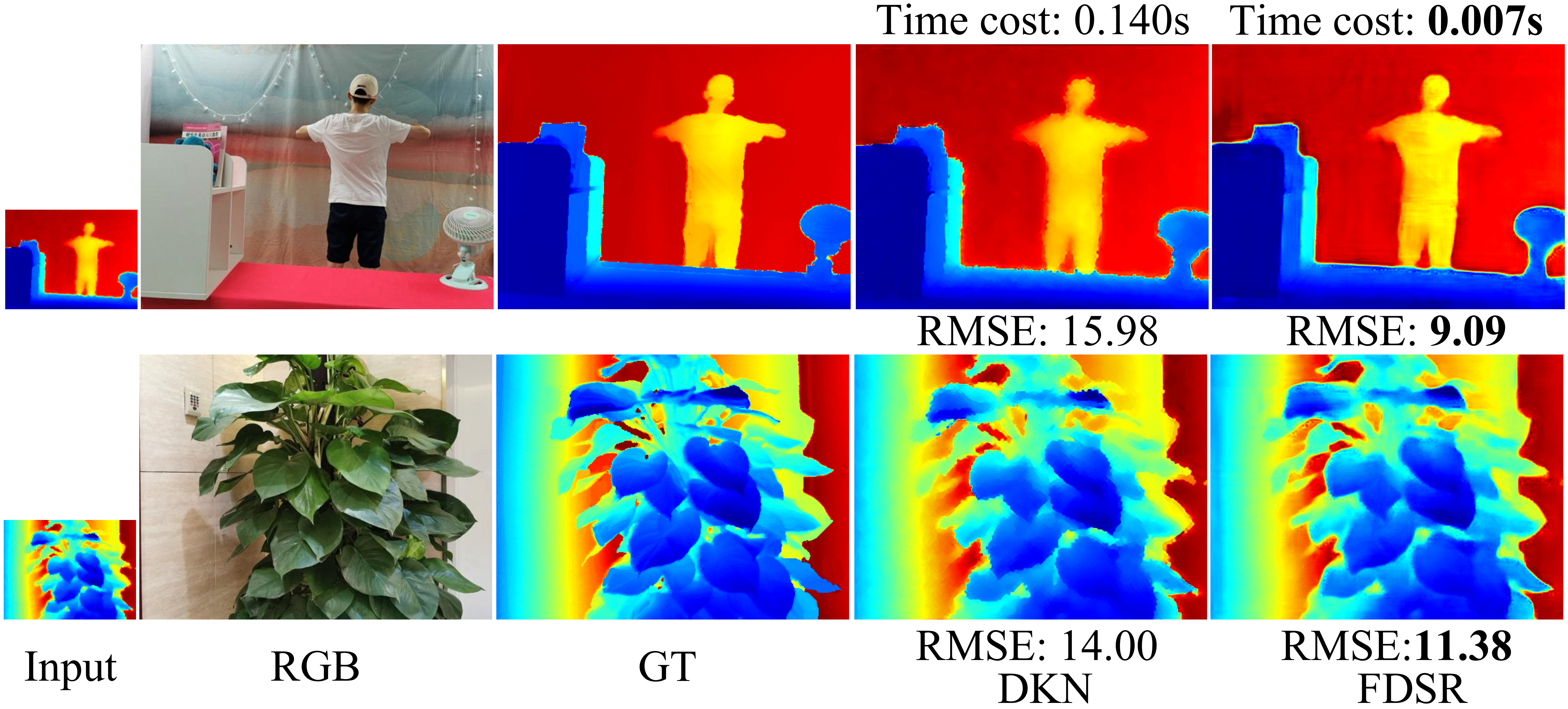}
	\end{center}
	\caption{RGB-D-D dataset display and depth map SR results comparison. Depth map SR results given by DKN~\cite{kim2020deformable} and FDSR is shown in the last two columns. The quantitive results in terms of RMSE is shown blow each row (lower is better). The runing time of DKN~\cite{kim2020deformable} and FDSR is shown on the top of the figure. }
	\label{fig:show}
\end{figure}

The existing ``RGB-D'' depth map SR datasets~\cite{song2015sun,butler2012naturalistic,hirschmuller2007evaluation} mainly focus on using a single HR depth map to generate paired LR and HR depth map correspondences through the downsampling strategy. In the real applications, the depth map SR task is more challenging and complicated because the real LR depth maps captured by depth sensors generally contain some noise even depth holes. Therefore, for the real scenes and real correspondences, we construct a large-scale paired depth map SR dataset named ``RGB-D-D'', which includes 4811 paired samples ranging from indoor scenes to challenging outdoor scenes. The ``D-D'' in our dataset represents the paired LR and HR depth maps captured from the mobile phone (LR sensors) and Lucid Helios (HR sensors)~\cite{Helios}, respectively. The dataset can offer two LR depth maps as input to evaluate the depth map SR task: LR depth map downsampled from the HR ground truth like previous research, and the raw LR depth map captured by LR sensor facing the real application scene. 
Besides, our dataset can contribute to many popular application scenarios of mobile phone and other depth-related tasks, such as portrait photography~\cite{Luo_2020_CVPR}, object modeling~\cite{kolev2014turning}, depth estimation~\cite{lee2019big}, depth completion~\cite{zhang2018deep}, \etc

Although numerous algorithms have been proposed for depth map SR and presented impressive performance, there are still some unsatisfactory points in detail preserving, computation complexity, and real-world application. Firstly, the sharp boundaries and elaborate details in the depth map SR are hard to recover especially when the scaling factor is large. Therefore, color image guided methods~\cite{xie2015edge,yu2015multi} are introduced to solve this problem. Different from them, we design a high-frequency guided multi-scale dilated structure to introduce the color guidance in an image decomposition manner and exploit the contextual information under different receptive fields. Secondly, to be applied on the platforms of the mobile devices and embedded systems, the depth map SR algorithms should take into account both the efficiency and accuracy. Inspired by~\cite{chen2019drop}, we design a high-frequency layer in our network, where the high-frequency features from RGB image are only used as the guidance in the early stages of depth map reconstruction branch, and the low-frequency components are suppressed to reduce the parameters. Lastly, the existing methods use downsample operation to get paired HR and LR depth maps for training which fails to simulate the real correspondences between HR and LR depth maps. We use the paired depth maps in RGB-D-D dataset to train a model, which greatly improve the value accuracy and visual effects in the real-world depth map SR task.

Focused on the real-world applications and the practical demands, we construct a large-scale and real-world depth map SR benchmark dataset and provide a fast solution for the depth map SR task. The contributions are highlighted in the following aspects:

\begin{itemize}
	\item We build the first and large-scale depth map SR benchmark dataset named RGB-D-D dataset\footnote{Refer to \url{http://mepro.bjtu.edu.cn/resource.html} for the RGB-D-D dataset download link.}, towards the real scenes and real correspondences. This dataset bridges the gap between theoretical research and real-world applications, and also flourishes the depth-related tasks in terms of benchmark dataset.
	
	\item We design a fast depth map super-resolution (FDSR) baseline, in which a high-frequency guided multi-scale structure is introduced to provide the frequency guidance and exploit the contextual information. Such decomposition strategy can improve the efficiency while retaining the reconstruction performance.
	\item Our network achieves the superior performance on the public datasets and our RGB-D-D benchmark dataset in terms of the speed and accuracy. Moreover, for the real-world depth map SR task, our algorithm can generate more accurate results with clearer boundaries and to some extent correct the value errors.
	
\end{itemize}

\section{Related Work}
In this section, we will briefly introduce the related benchmark datasets and algorithms for depth map SR.

\vspace*{2mm}\noindent {\bf Benchmark Datasets.}
There are various RGB-D datasets used for training and evaluating the depth map SR task. These datasets can be roughly divided into the synthetic datasets and the real-scene datasets. The synthetic datasets are built by synthetic computer graphic techniques and offer relatively high-quality data, such as New Tsukuba~\cite{peris2012towards}, Sintel~\cite{butler2012naturalistic} and ICL~\cite{handa2014benchmark}. Limited by the discrepancy of virtual scenes and real scenes, some datasets towards real scene are constructed. Middlebury dataset~\cite{scharstein2003high,scharstein2007learning,hirschmuller2007evaluation,scharstein2014high} provides a few samples containing high-quality and noise-free depth maps.
Focusing on the indoor scenes, NYU v2 dataset~\cite{Silberman:ECCV12} and SUN RGBD dataset~\cite{song2015sun} are built, where NYU v2~\cite{Silberman:ECCV12} includes 1449 RGB-D pairs, and SUN RGBD dataset~\cite{song2015sun} consists of three different RGB-D datasets (\ie, NYU v2~\cite{Silberman:ECCV12}, B3DO~\cite{janoch2013category} and SUN3D~\cite{xiao2013sun3d}).
However, the mentioned datasets only face to the real scenes but fail to build the real correspondences between HR and LR depth maps, which are important for real-world depth map SR. To this end, we construct the first real-world depth map SR dataset, which not only faces the real scenes in practical applications, but also meets the real correspondences of LR and HR depth maps.
\begin{figure*}[!htbp] 
	\input{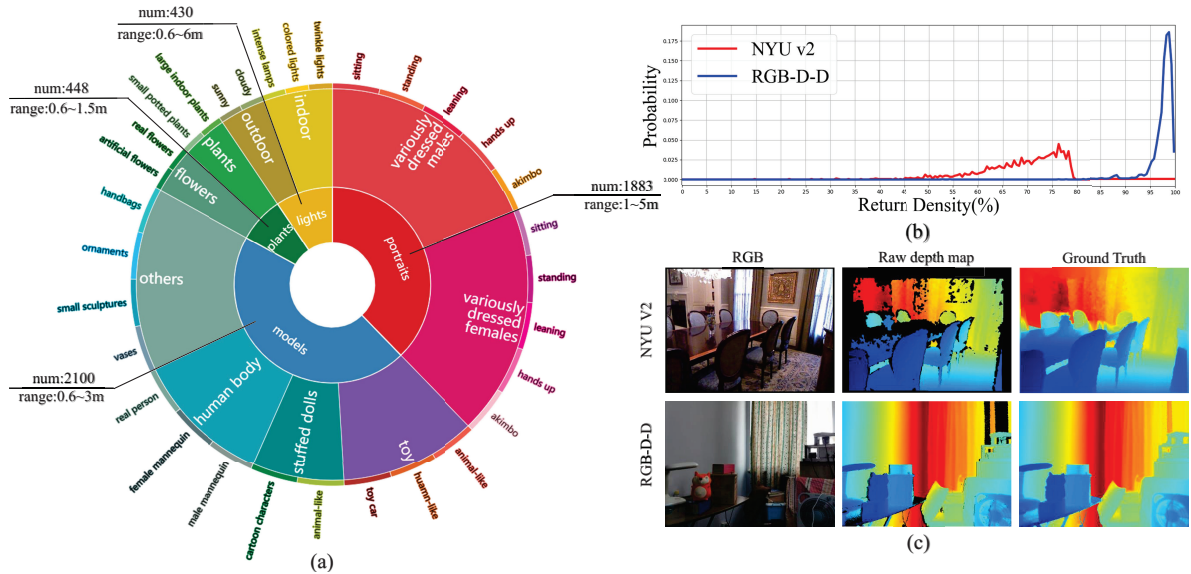}
	\centering
	\caption{\textbf{Dataset statistic.} (a) The scenes and corresponding hierarchical content structure of RGB-D-D. The inner ring represents the classification of scenes. The legends express the number of samples and the depth range of corresponding subsets. (b) The fitted probability density curves of return density for the raw depth maps from NYU v2~\cite{Silberman:ECCV12} and RGB-D-D. (c) Examples of the raw depth map and ground truth from NYU v2~\cite{Silberman:ECCV12} and RGB-D-D. Black region indicates the missing depth value.}
	\label{fig:dataset}
\end{figure*}

\vspace*{2mm}\noindent {\bf Algorithm Models.}
According to the characteristics of input data, depth map SR algorithms can be categorized into two categories: non RGB-guided depth map SR and RGB-guided depth map SR. Non RGB-guided methods~\cite{riegler2016atgv,xie2015edge} only used LR depth maps as input to produce HR ones. These methods do not fully utilize the color information that may induce unsatisfying performance. By contrast, RGB-guided methods~\cite{kim2020deformable,li2019joint,wen2018deep,hui2016depth} have become the mainstream of this task. For the unsupervised methods~\cite{park2011high,yang2012depth,ferstl2013image,li2012joint}, the depth map SR task is generally modeled as an optimization problem. As for the learning-based methods, the RGB information is used as features directly or used to convert and produce other types of guidance information. In~\cite{kwon2015data,hui2016depth,wen2018deep,zuo2019multi,kim2020deformable}, the authors extract features of multi-level from RGB image to solve the depth map SR task. While~\cite{li2016deep,pan2019spatially,li2019joint} concern about the RGB images and try to use the HR RGB images to produce more useful features which achieves better performance. However these methods only put more efforts on the accuracy improvement which may lead to high computation complexity. Thus, we propose a fast depth map SR method to well balance efficiency and accuracy. Though, there are some works\cite{guo2018hierarchical} use the real-scene LR depth maps to evaluate their methods, they fail to solve the problem at source. We are the first to use the paired depth maps in RGB-D-D dataset to train and simulate the correspondences which greatly improve the effectiveness of depth map SR.

\section{RGB-D-D Dataset}
\label{section:RGBDD}
We collect the first real-world depth map SR dataset which contains a total of 4811 RGB-D-D pairs. Each image pair contains the HR color images from mobile phone, the real-world LR depth maps captured by the low-power Time of Flight (ToF) camera on mobile phone, and the HR depth maps captured by industrial ToF camera.

\subsection{Dataset Collection}
\vspace*{2mm}\noindent {\bf Acquisition Devices.}
We use the Huawei P30 Pro to collect color images and LR depth maps. The Huawei P30 Pro has a 40 million pixels Quad RYYB sensor which can capture  $3648 \times 2736$ HR color image, and a ToF camera with $240 \times 180$ resolution. The HR depth maps are captured by Helios ToF camera~\cite{Helios} produced by LUCID vision labs. 
They use the same depth acquisition principle which ensures the depth values captured by them are almost the same.
Meanwhile, we guarantee little missing values of LR depth maps by limiting the farthest distance of backgrounds.

\vspace*{2mm}\noindent {\bf Data Processing.}
We calibrate the primary camera of the phone with the Helios ToF camera, and align them on the $640 \times 480$ resolution color image by the intrinsic and extrinsic parameters.  
Due to the different field of view (FOV) between them, the $640 \times 480$ raw point cloud of Helios is projected on the center area of the corresponding $640 \times 480$ resolution color image, and finally generate a dense and high-quality depth map which is smaller than $640 \times 480$. Then we crop it as the 512 $\times$ 384 HR depth map, which corresponds to the central $192 \times 144$ area of the LR depth map with the same scale variations.
Towards the depth holes caused by the occlusion effect of projection processing and some low-reflection objects (such as glass surface and infrared absorbing surface), we firstly use the over-segmentation algorithm~\cite{nguyen2003watersnakes} to get plentiful boundary information of color image. And the colorization method \cite{levin2004colorization} is used to fill holes in the supervision of boundary information. Clear depth edges can be filled according to the obvious border between foreground and background of color image, especially for strip-shaped holes on background caused by the occlusion effect of projection processing.

\vspace*{2mm}\noindent {\bf User Study.}
The four-round different people evaluation scores the filled HR depth maps to judge whether they could be the ground truth (the full mark of every round is 10). After four-round evaluation, the filled depth maps with total scores beyond 35 become the part of ground truth samples. Besides, some of the eliminated depth maps have high evaluation scores (beyond 30) and few repairable defects. We further introduce the manual scribbles~\cite{wang2011stereobrush} on them to get convincing depth maps by means of user intervention. After the additional four-round blind selection, these remaining depth maps which achieve good visual effects on edge serve as the other part of ground truth samples.

\subsection{Dataset Statistic}
\vspace*{2mm}\noindent {\bf Real Scenes.}
We collect the paired ``RGB-D-D'' samples in various scenes as shown in Figure \ref{fig:dataset} (a). The RGB-D-D dataset is divided into four main categories: portraits, models, plants, and lights.
The portraits category is mainly for the real applications of depth-of-field blur~\cite{Luo_2020_CVPR} in portrait photography. The traditional stereo camera cannot acquire satisfying depth information to simulate the large aperture in the repeated or weak texture scenes, so we collect plentiful samples containing the human body with pose variation as the foreground in different backgrounds.
The models category can be used to optimize the edge of objects in depth maps when modeling the object by LR depth maps in sequential views~\cite{marchand2015pose}. We collect different depth maps of the relatively static objects on a rotating booth by discretely extracting video frames. 
The plants category which has dense branches and leaves contains a lot of structural and hierarchical details that LR depth sensors cannot capture. It is a challenge to the depth enhancement algorithms to infer these details by low-quality depth map and RGB. So we capture images containing various kinds of luxuriant plants and flowers in close range.
In addition, strong indoor light sources and outdoor light have a great impact on the quality of depth maps, especially for the low-power sensors. Therefore, the lights category can be used to improve the quality of depth maps as close to the high-performance camera as possible, and to explore the effect of complex illumination environment to the color guided depth SR algorithms.

\vspace*{2mm}\noindent {\bf Real Correspondences.}
Actually there are only LR depth maps when applying the depth map SR algorithm in real-world applications. The relationship between LR and HR depth maps can not be simulated by traditional downsampling operations. Hence, it is necessary to capture the paired LR and HR depth maps by devices with different resolution. Both the LR depth map and the color image captured by the phone have aligned in $192 \times 144$ resolution. Meanwhile, the projected HR depth map from Helios~\cite{Helios} also align with color image in $512 \times 384$ resolution. The alignment of these three depth maps at different scales guarantee the real correspondences of LR and HR depth map.

\vspace*{2mm}\noindent {\bf High Quality.}
Because of the inevitable error values in the filled depth maps, the quality of preprocessed depth maps often degrade. Facing the practical applications, we capture the depth maps under the suggested distance to obtain depth maps with low rate of missing values and clear boundaries. Besides, the existence of many-to-one relationship in the process of raw point cloud projects to the image plane ensures that our raw depth maps contain dense depth values. All these can guarantee the high quality of our raw depth map. Figure \ref{fig:dataset} (b) shows the comparison between RGB-D-D and NYU v2~\cite{Silberman:ECCV12} on the statistic distribution of return density of which most of our raw HR depth maps are more than $90\%$. As shown in Figure \ref{fig:dataset} (c), benefiting by the higher return density, our ground truth has less depth values errors and better boundaries than NYU v2~\cite{Silberman:ECCV12}.

\begin{figure*}[!htbp]
	\input{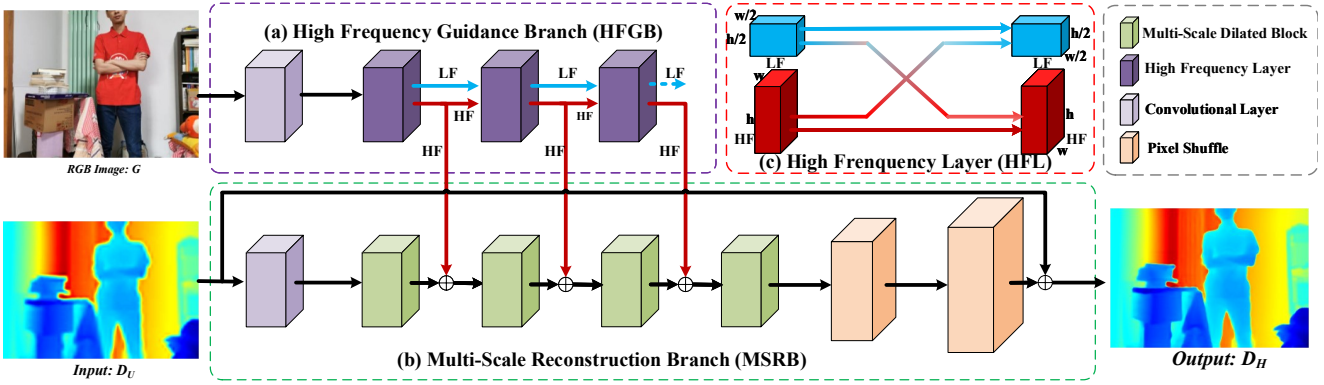}
	\centering
	\caption{Overview of FDSR architecture. MSRB uses input depth map, high-frequency components extracted from RGB image to generate HR depth map. The blue lines and red lines in (a) indicate the low- and high-frequency components splited by HFL, respectively. (c) shows how the HFL works.  }
	\label{Fig:FDSR Archetecture}
\end{figure*}

\section{Proposed Framework}
\subsection{Problem Formulation}
Given a LR depth map $D_{L}\in\mathbb{R}^{M\times N\times1}$ and the corresponding HR RGB image $G\in\mathbb{R}^{sM\times sN\times3}$, the purpose of this work is to recover a HR depth map $D_{H}\in\mathbb{R}^{sM\times sN\times3}$ with the guidance of $G$, where $M$ and $N$ denote the height and width of $D_{L}$, respectively, $s$ is the scaling factor. We use bicubic interpolation to upscale $D_{L}$ to HR space, which results in $D_{U}\in\mathbb{R}^{sM\times sN\times1}$. As shown in Figure~\ref{Fig:FDSR Archetecture}, we feed the paired $D_{U}$ and $G$ into our network to learn the non-linear mapping from $D_{U}$ to $D_{H}$ through residual learning. Such process can be formulated as 
\begin{equation}
	D_{H} = D_{U} + \mathcal{F}(D_{U}, \mathcal{H}(G); \theta)
	\label{eq:1}
\end{equation}
where $\mathcal{F}(\cdot)$ is a function to learn the residual mapping between $D_{U}$ and $D_{H}$, the $G$ is embedded into a high-frequency extractor $\mathcal{H}(\cdot)$ to provide high-frequency guidance for depth map SR, and $\theta$ is the learned weights set. 

\subsection{Network Architecture}
Figure \ref{Fig:FDSR Archetecture} outlines the whole architecture of our fast depth super-resolution network called FDSR, which consists of a high-frequency guidance branch (HFGB) and a multi-scale reconstruction branch (MSRB). Our framework progressively equip with four multi-scale reconstruction blocks to exploit the contextual information under different receptive fields in MSRB, meanwhile, the high-frequency guidance extracted from the HFGB is integrated with the multi-scale contextual information to enhance the ability of detail recovery for depth map SR. Finally, the comprehensive and discriminative reconstruction features are fed into a residual mapping function to generate HR depth map.

\vspace*{2mm}\noindent {\bf High-Frequency Guidance Branch.}
Motivated by previous methods~\cite{li2019joint,zuo2019multi}, we design a high-frequency layer (HFL) to adaptively highlight the high-frequency components and suppress the low-frequency component. Different from existing methods, we put more efforts on the following two aspects (1) a direct high-frequency decomposition method is designed, where the octave convolution~\cite{chen2019drop} is utilized to decompose the RGB features into high- and low-frequency components. (2) the high-frequency components are effectively used to guide depth map SR. Such design focuses on the useful high-frequency detail information to improve the performance, while it reduces the computation complexity due to the low-frequency components are not used in the MSRB. As shown in the Figure \ref{Fig:FDSR Archetecture} (c), the previous high- and low-frequency feartures are embedded into the HFL to generate the current ones, which can be formulated as follows:

\begin{equation}
	\begin{split}
		Y^{H}_{i+1} = f(Y^{H}_{i}; W^{H \rightarrow H}_{i}) + up(f(Y^{L}_{i}; W^{L \rightarrow H}_{i}), 2) \\
		Y^{L}_{i+1} = f(Y^{L}_{i}; W^{L \rightarrow L}_{i}) + f(down(Y^{H}_{i},2); W^{H \rightarrow L}_{i})
	\end{split}
	\label{eq:2}
\end{equation}
where $Y^{H}_{i+1}$ and  $Y^{L}_{i+1}$ denotes the high- and low-frequency features, respectively, the $W$ is convolutional kernel, $f(Y; W)$ is a convolutional operation for $Y$ with the kernel $W$, $up(Y, 2)$ is an upsample operation by a factor of $2$ via nearest interpolation, and $down(Y, 2)$ represents 2 $\times$ downsampling for Y by using average pooling operation.

\vspace*{2mm}\noindent {\bf Multi-Scale Reconstruction Branch.}
This branch aims to progressively recover HR depth map through utilizing mulit-scale contextual information. We first use one $3 \times 3$ convolution layer to initial feature extraction. Then, to exploit the contextual information under different receptive fields, we combine two dilated convolutions to form a multi-scale dilated block (MSDB), and one convolution layer is used to integrate the concatenated features:
\begin{equation}
	M_{i}(F_{i}) = W^{j}_{i}(\sum_{j\in K}(W_{M_{i}}^j \ast F_{i} + b_{M_{i}}^j))+ b_{i}^j
	\label{eq:3}
\end{equation}
where  $M_{i}$ is the $ith$ multi-scale block, $F_{i}$ is the input of $M_{i}$, $K = \{1,2\}$, $\ast$ denotes dilated convolution operation, $W$ and $b$ are parameters which the convolution layer should learn. Our MSDB not only enlarges the receptive field, but also enriches the diversity of convolutions, which results in an ensemble of convolutions with different receptive regions and dilation rates.

As for feature combination, three levels of high-frequency features extracted by HLFs are fused with different MSDBs respectively in the early stage of MSRB. What we have to emphasize is that, we split the output of each HLF and only use the extracted high-frequency component as guidance. Each stage of feature fusion $F_{add}^{i}$  can be described as follows:
\begin{equation}
	F^{i + 1} = Y^{H}_{i} \oplus M_{i}(F_{i})
	\label{eq:4}
\end{equation}
where $i = {1, 2, 3}$, and $\oplus$ is concat operation. Then, the final fusion features $ F^{4}$ is embedded into the last MSDB to generate reconstruction features.

In order to better apply FDSR in the platform of mobile devices and embedded systems, we mainly use two key operations to make our network faster. First, the designed HFL adaptively extracts the high-frequency features what we should focus on. Therefore, the parameters are reduced proportionately while ensuring effective performance. Second, in the stage of data preparation, we resample the input data by transform the color image to gray scale. Then we split the gray image and the input depth map to $r^{2}$ pieces of blocks of size $ h/r \times w/r$ and stack them together.  In the end, we use pixel shuffle operation to recover the HR depth map to the original size of $h \times w \times 1$. 

\begin{figure*}[!htbp]
	\input{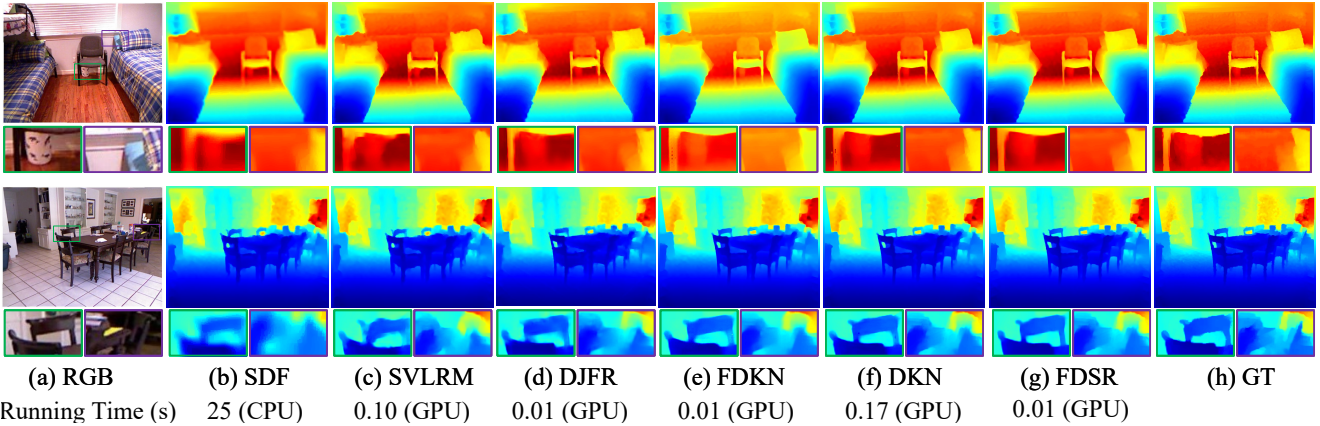}
	\centering
	\caption{Visual comparison of $\times 8$ depth map SR results on NYU v2~\cite{Silberman:ECCV12}. (a) RGB images. (b) SDF~\cite{li2016deep}. (c) SVLRM~\cite{pan2019spatially}. (d) DJFR~\cite{li2019joint}. (e) FDKN~\cite{kim2020deformable}. (f) DKN~\cite{kim2020deformable}. (g) FDSR (trained on NYU v2). (h) GT. The GPU time is tested on a NVIDIA GTX TITAN XP GPU.}
	\label{Fig:fig1}
\end{figure*}

\begin{table*}[!htbp] 
	\centering	
\small
\resizebox{\textwidth}{!}{
\begin{tabular}{l|ccccccccccccccc}
	\toprule
	RMSE & Bicubic & MRF~\cite{diebel2006application}   & GF~\cite{2010Guided}    & JBU~\cite{2007Joint}   & TGV~\cite{2013Image}   & Park~\cite{park2011high} & SDF~\cite{li2016deep}   & FBS~\cite{2015The}   & DMSG~\cite{hui2016depth} & PAC~\cite{2019Pixel}  & DJF~\cite{li2016deep}   & DJFR~\cite{li2019joint}  & DKN~\cite{kim2020deformable} &FDKN~\cite{kim2020deformable} & FDSR   \\ \midrule
	$\times 4$   & 8.16    & 7.84  & 7.32  & 4.07  & 4.98  & 5.21 & 5.27  & 4.29  & 3.02 & 2.39 & 3.54  & 3.38  & 1.62 & 1.86&\textbf{1.61}        \\ 
	$\times 8$   & 14.22   & 13.98 & 13.62 & 8.29  & 11.23 & 9.56 & 12.31 & 8.94  & 5.38 & 4.59 & 6.2   & 5.86  & 3.26 & 3.58&\textbf{3.18}       \\ 
	$\times 16$  & 22.32   & 22.2  & 22.03 & 13.35 & 28.13 & 18.1 & 19.24 & 14.59 & 9.17 & 8.09 & 10.21 & 10.11 & 6.51 & 6.96&\textbf{5.86}     \\ \bottomrule
\end{tabular}	
}
	\caption{Comparisons with the state-of-the-art methods in terms of RMSE on NYU v2~\cite{Silberman:ECCV12}. The depth values are measured in centimeter.}
	\label{tab:NYU_rmse}
\end{table*}

\subsection{Loss Function}
We train our model by minimizing the $L_{1}$ norm between the output of our method $\mathcal{F}(\cdot)$ and ground truth as follows:

\begin{equation}
	L(\hat{\mathcal{F}},\mathcal{F}^{gt}) = \sum_{P} \lVert \mathcal{F}_{p}^{gt} - \hat{\mathcal{F}_{p}} \rVert_{1}
	\label{eq:5}
\end{equation}
where $\hat{\mathcal{F}}$ and $\mathcal{F}^{gt}$ denote the depth SR result and ground truth, respectively, $\lVert \cdot \rVert_{1}$ computes the $L_{1}$ norm, $P$ is the set of all pixels and $p$ represents a pixel in an image.

\section{Experiments}
\subsection{Datasets and Implementation Details}
To evaluate the performance of different methods, we conduct sufficient experiments on the public NYU v2 dataset~\cite{Silberman:ECCV12} to and our real-world RGB-D-D dataset.

As for public dataset, we choose the widely used depth map SR dataset NYU v2~\cite{Silberman:ECCV12}, and evaluate ours and other methods on it. Following~\cite{kim2020deformable}, we sample 1000 RGB-D image pairs of size $640 \times 480$ from the NYU v2 dataset for training and the rest 449 image pairs for testing. As for RGB-D-D dataset, we randomly split 1586 portraits, 380 plants, 249 models for training and 297 portraits, 68 plants, 40 models for testing. Our FDSR is implemented in PyTorch on a PC with an NVIDIA GTX TITAN XP GPU. A MindSpore implementation version is also provided. Limited by the length of the paper, more details of experimental settings can be found in the supplemental materials. 


\subsection{Experiments on NYU v2 Dataset}

\begin{table}[!htbp]
	\centering
\small
\begin{tabular}{cccc|ccc} 
\toprule
\multirow{2}{*}{Percentage} & \multicolumn{3}{c |}{Value Errors (in 10 m)} & \multicolumn{3}{c}{Edge Errors}                  \\ \cmidrule{2-7} 

                              & $\times 4$   & $\times 8$  & $\times 16$   &$\times 4$  &$\times 8$  & $\times 16$          \\ \midrule
                            
SDF~\cite{li2016deep}                       &0.42&1.28&3.52&4.20&10.19&25.06 \\ 
SVLRM~\cite{pan2019spatially}       &1.08&2.56&5.76&6.04&24.28&49.26                     \\
DJF~\cite{li2016deep}       &1.05&2.74&6.25&9.87&30.38&55.35 \\ 
DJFR~\cite{li2019joint}      &1.04&2.72&6.25&6.78&25.01&53.98  \\ 
FDKN~\cite{kim2020deformable}    &0.04&0.24&1.00&0.83&3.27&13.03 \\
DKN~\cite{kim2020deformable}  &0.05&0.20&1.10&0.95&2.95&13.78   \\ 
FDSR                       &\textbf{0.04}&\textbf{0.18}&\textbf{0.69}&\textbf{0.78}&\textbf{2.60}&\textbf{9.44}  \\ 

\bottomrule
\end{tabular}

	\caption{Value errors and edge errors on NYU v2~\cite{Silberman:ECCV12}.}
	\label{tab:error_nyu}
\end{table}
\begin{figure*}[!htbp]
	\input{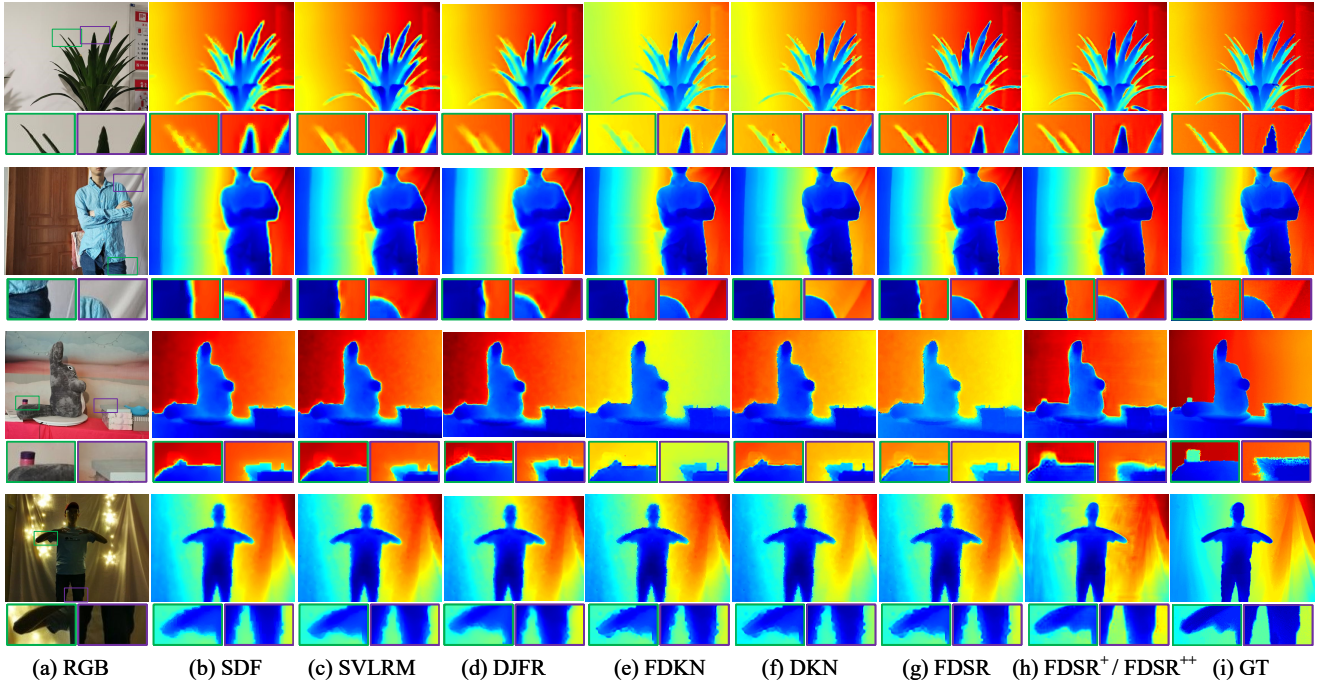}
	\centering
	\caption{Visual comparison of $\times 8$ depth map SR results on RGB-D-D. The first two and last two rows are the results of FDSR$^{+}$ and FDSR$^{++}$ respectively (a) RGB images. (b) SDF~\cite{li2016deep}. (c) SVLRM~\cite{pan2019spatially}. (d) DJFR~\cite{li2019joint}. (e) FDKN~\cite{kim2020deformable}. (f) DKN~\cite{kim2020deformable}. (g) FDSR (trained on NYU v2~\cite{Silberman:ECCV12}). (h) FDSR$^{+}$ / FDSR$^{++}$ (Trained in downsampling manner / Trained in real-world manner on RGB-D-D). (i) GT.}
	\label{Fig:fig23}
\end{figure*}
\begin{table*}[!htbp]
	\centering	
\small
\begin{tabular}{l|cccccccccc}
\toprule
RMSE  & \multicolumn{1}{c}{SDF~\cite{li2016deep}} & \multicolumn{1}{c}{SVLRM~\cite{pan2019spatially}} & \multicolumn{1}{c}{DJF~\cite{li2016deep}} & 
\multicolumn{1}{c}{DJFR~\cite{li2019joint}} & \multicolumn{1}{c}{FDKN~\cite{kim2020deformable}} & \multicolumn{1}{c}{DKN~\cite{kim2020deformable}} &
 \multicolumn{1}{c}{FDSR} &
  \multicolumn{1}{c}{FDSR$^{+}$}  \\ \midrule
$\times 4$  & 2.00& 3.39&3.41 &3.35 &1.18& 1.30 &1.16 & \textbf{1.11}                   \\ 
$\times 8$  &3.23 & 5.59& 5.57 &5.57 &1.91& 1.96 & 1.82 & \textbf{1.71}               \\ 
$\times 16$ &5.16 &8.28 &  8.15 &7.99 & 3.41& 3.42 & 3.06 & \textbf{3.01}  \\ 
\bottomrule
\end{tabular}

	\caption{Quantitative depth map SR results on RGB-D-D. FDSR$^{+}$ is trained in downsampling manner on RGB-D-D)}
	\label{tab:MeproD_experiment}
\end{table*}
As for training on NYU v2 dataset~\cite{Silberman:ECCV12}, we obtain the LR depth maps from ground truth by using bicubic downsampling operation. The initial learning rate is 0.0005 and reduce to half every 80k iterators and the training is stopped after 100 epochs since more epochs do not provide more improvement. We compare our FDSR with other methods with the scaling factors of 4, 8, 16. The quantitative results are shown in Table~\ref{tab:NYU_rmse}. It can be observed that our method achieves the best performance on NYU v2.

To further analyze the robustness of our method, we conduct two extra experiments on NYU v2: (1) depth value errors to inflect the global depth map SR accuracy, (2) edge errors to measure the local accuracy. We report the value errors which is calculated by the percentage of value errors over 10\% between ground truth and output. As for edge errors, we report the percentage of errors over $1.2\%$ in the edge area. The details of calculation process for value errors and edge errors will be described in supplement material.
Observing Table~\ref{tab:error_nyu}, FDSR has both less value errors and edge errors, which means our method produces more accurate results globally and locally. 

As for qualitative results, we show the visual comparison for $\times 8$ depth map SR in Figure \ref{Fig:fig1}. The overall and details of the results demonstrate that the proposed method FDSR can obtain more accurate depth map values. Our results show finer boundaries and more visual pleasant details without the texture-copy artifacts and extra noise introduced .

The running time is also shown in Figure~\ref{Fig:fig1}. The size of input is $640 \times 480$. Our FDSR method achieves the comparable efficiency with DJFR~\cite{li2019joint} and FDKN~\cite{kim2020deformable} while the performance of FDSR is better than anyone of them.

\begin{table*}[!htbp]
	\centering	
\small
\begin{tabular}{l|cccccccccc}
\toprule
&  \multicolumn{1}{c}{SDF~\cite{li2016deep}} & \multicolumn{1}{c}{SVLRM~\cite{pan2019spatially}} & \multicolumn{1}{c}{DJF~\cite{li2016deep}  } & \multicolumn{1}{c}{DJFR~\cite{li2019joint} } & \multicolumn{1}{c}{FDKN~\cite{kim2020deformable} } & \multicolumn{1}{c}{DKN~\cite{kim2020deformable} } &  \multicolumn{1}{c}{FDSR} & \multicolumn{1}{c}{FDSR$^{++}$}  \\ \midrule
 RMSE           &7.16 & 8.05& 7.90 &8.01 & 7.50&  7.38 & 7.50  &  \textbf{5.49}              \\ 
 Value Errors    &2.86 & 3.62& 3.62 &3.67 &2.85 & 2.83  & 2.90  &    \textbf{1.71}            \\ 
 Edge Errors    &52.78 & 51.87&50.56 &52.28 &51.73 & 51.90  &  51.89 & \textbf{42.89}               \\ 
\bottomrule
\end{tabular}

	\caption{RMSE, value errors and edge errors of depth SR results. FDSR$^{++}$ is trained on RGB-D-D in real-world training manner.}
	\label{tab:phone_rmse}
\end{table*}

\subsection{Experiments on RGB-D-D Dataset}
To verify the generalizability of RGB-D-D, we conduct sufficient experiments on it. The experiments on our dataset also further demonstrate the performance of our algorithm. 

\vspace*{2mm}\noindent {\bf Testing without Retraining on RGB-D-D.}
Firstly, to make a fair comparison with other algorithms on our dataset, we conduct experiments among models trained on NYU v2~\cite{Silberman:ECCV12} without retraining. The quantitative results in terms of RMSE are shown in Table~\ref{tab:MeproD_experiment}. The value errors and edge errors on RGB-D-D given by each algorithm are also reported in Table~\ref{tab:error_ours}. The smaller RMSE value, value errors and edge errors of FDSR among evaluated methods demonstrate the accuracy and effectiveness of our algorithm. Figure~\ref{Fig:fig23} illustrates the qualitative results. The evaluated methods, SDF~\cite{li2016deep}, SVLRM~\cite{pan2019spatially} and DJFR~\cite{li2019joint} cannot recover clear boundaries and fine details. Though DKN~\cite{kim2020deformable} and FDKN~\cite{kim2020deformable} produce clear boundaries, they have larger global errors and even some noises are brought in.

\begin{table}[!htb]
	\centering
\small
\begin{tabular}{cccc|ccc} 
\toprule
\multirow{2}{*}{Percentage} & \multicolumn{3}{c |}{Value Errors (in 3 m)} & \multicolumn{3}{c}{Edge Errors}                  \\ \cmidrule{2-7} 

                              & $\times 4$   & $\times 8$  & $\times 16$   &$\times 4$  &$\times 8$  & $\times 16$         \\ \midrule

SDF~\cite{li2016deep}                         &0.33&0.90&2.37&3.22&8.74&20.71\\ 
SVLRM~\cite{pan2019spatially}                      &0.80&2.11&4.58&5.08&15.18&34.30                     \\
DJF~\cite{li2016deep}                        &0.82&2.19&4.89&5.65&17.07&35.32 \\ 
DJFR~\cite{li2019joint}                        &0.79&2.15&4.78&5.26&15.66&34.54  \\ 
FDKN~\cite{kim2020deformable}                       &0.11&0.28&0.94&1.39&3.41&11.73 \\
DKN~\cite{kim2020deformable}                        &0.14&0.33&1.54&2.11&3.55&12.93   \\ 
FDSR                       &0.10&0.26&0.76&1.38&3.09&12.47 \\ 
FDSR$^{+}$                       &\textbf{0.09}&\textbf{0.21}&\textbf{0.67}&\textbf{1.15}&\textbf{2.79}&\textbf{11.68}  \\ 

\bottomrule
\end{tabular}

	\caption{Value errors and edge errors of depth SR results on RGB-D-D. FDSR$^{+}$ is trained in downsampling training manner.}
	\label{tab:error_ours}
\end{table}

\vspace*{2mm}\noindent {\bf Training in Downsampling Manner on RGB-D-D.}
We retrain our models on the training set of RGB-D-D dataset to demonstrate the effectiveness of our dataset and our model. We use downsampled LR depth maps as input. When training on RGB-D-D dataset, the initial learning rate is 0.0005 and reduce to half every 40k iterators and the training for each scaling factor model is stopped after 40 epochs. 

The results are appended in Table \ref{tab:MeproD_experiment}, Table \ref{tab:error_ours} and Figure~\ref{Fig:fig23}, which are obviously improved by training on our training data. Benefiting by the more clearly and sharper boundaries in our training and testing data, our model can achieve better performance, especially on the boundaries of objects and accuracy of depth values.


\vspace*{2mm}\noindent {\bf Training in Real-World Manner on RGB-D-D.}
To make full use of our proposed RGB-D-D dataset, we train our model on the training set by utilizing the LR depth maps as input. Before evaluating, the missing holes of raw LR depth maps are filled by the colorization method \cite{levin2004colorization}. The size of the LR depth map is $192 \times 144$ and the target resolution is $512 \times 384$. The settings and training strategies are as same as we trained on HR depth maps of RGB-D-D.
We test all the evaluated methods on the filled LR depth maps via using the existing $\times 4$ models and the results can be seen in Table~\ref{tab:phone_rmse}. We append our results obtained by the model trained on the paired LR and HR depth maps. It can be observed that, all the evaluated methods have bad performance facing the real-world depth map SR task, which means the traditional downsample training strategy fails to model the real-correspondence between LR and HR depth maps. Observing the last two rows of Figure~\ref{Fig:fig23}, after retraining FDSR on the paired LR and HR depth map in RGB-D-D dataset, the visual effects and value accuracy are greatly improved, which demonstrates that our dataset reflects the real-correspondences characteristics between LR and HR depth maps. Thus, the RGB-D-D dataset has great potential to promote the development of real-world depth map SR.

We also conduct experiments on the group of lights in our RGB-D-D dataset. It is a very challenging set of data, because the illumination intensity is complicated and the LR depth maps are in lower quality with bigger missing holes. Observing the last two rows in Figure~\ref{Fig:fig23}, we obtain a better result with good boundaries, more accuracy depth values and more pleasant visual effects, while other algorithms fail to recover good HR depth maps. 

\subsection{Ablation Study}

To demonstrate the effectiveness of the designed architecture of our depth map SR baseline, we conduct serveral ablation studies. For such an ablation study, the basic setup refers to the experiments above. The results in Table~\ref{tab:ablation} clearly demonstrates that both the HFL and HFGB can be used to improve the performance of FDSR. What's more, the improvement of FDSR implies that the employing HFL components that HFGB to great extent.
\begin{table}[htbp]
	\centering
\small
\begin{tabular}{cccc|ccc} 
	\toprule
	\multirow{2}{*}{Methods} & \multicolumn{3}{c |}{NYU v2~\cite{Silberman:ECCV12}} & \multicolumn{3}{c}{RGB-D-D}                  \\ \cmidrule{2-7} 
	
	& $\times 4$   & $\times 8$  & $\times 16$   &$\times 4$  &$\times 8$  & $\times 16$          \\ \midrule
w/o HFGB    & 2.02 &3.90 &7.58  &1.16&1.88&3.47     \\ 
w/o HFL     & 1.68 & 3.21& 5.89 &1.13&1.85&3.20     \\ 
FDSR        & \textbf{1.61} &\textbf{3.18} & \textbf{5.86 }&\textbf{1.11}&\textbf{1.71}& \textbf{3.01}\\ 
	\bottomrule
\end{tabular}

		\caption{RMSE evaluation of HFL and HFGB.}
	\label{tab:ablation}
\end{table}

\section{Conclusion}
Towards the real-world depth map SR, we build the first benchmark dataset which satisfy both real scene and real coorespondence.
The dataset contains paired LR and HR depth maps in multiple scenarios, and contributes the completely new benchmark dataset for real-world depth map SR research. 
Furthermore, the ``RGB-D-D'' triples not only can complete the traditional depth-related tasks, such as depth estimation, depth completion, \etc but also have significant potential to promote the application of depth maps on portable intelligent electronics.
We also provide a fast and accurate depth map SR baseline adaptively focusing on the high-frequency components of the guidance and suppress the low-frequency components. 
Our algorithm achieves the competitive performance on public datasets and our proposed dataset, what's more, it has an ability to cope with the task of real-world depth map SR.

\vspace*{2mm}\noindent {\bf Acknowledgements:} This work was supported by the National Key Research and Development of China (No. 2018AAA0102100), the National Natural Science Foundation of China (No. U1936212, 61972028), the Beijing Natural Science Foundation (No. JQ20022), the Beijing Nova Program under Grant (No. Z201100006820016) and the CAAI-Huawei MindSpore Open Fund.

{\small
	\bibliographystyle{ieee_fullname}
	\bibliography{egbib}
}

\end{document}